\documentclass{article}
\usepackage{ijcai22}

\usepackage{times}
\usepackage{soul}
\usepackage{url}
\usepackage[hidelinks]{hyperref}
\usepackage[utf8]{inputenc}
\usepackage[small]{caption}

\usepackage{subcaption}

\usepackage{graphicx}
\usepackage{amsmath}
\usepackage{amssymb}
\usepackage{amsthm}
\usepackage{booktabs}
\usepackage{algorithm}
\usepackage{algorithmic}
\title{Explaining Imitation Learning through Frames}

\author{
Boyuan Zheng$^1$
\and
Jianlong Zhou$^1$\and
Chunjie Liu\and
Yiqiao Li$^1$\And
Fang Chen$^1$
\affiliations
$^1$University of Technology Sydney\\
\emails
Boyuan.Zheng-1@student.uts.edu.au,
Jianlong.Zhou@uts.edu.au
}

\begin{document}

\maketitle

\begin{abstract}
As one of the prevalent methods to achieve automation systems, Imitation Learning (IL) presents a promising performance in a wide range of domains. However, despite the considerable improvement in policy performance, the corresponding research on the explainability of IL models is still limited. Inspired by the recent approaches in explainable artificial intelligence methods, we proposed a model-agnostic explaining framework for IL models called R2RISE. R2RISE aims to explain the overall policy performance with respect to the frames in demonstrations. It iteratively retrains the black-box IL model from the randomized masked demonstrations and uses the conventional evaluation outcome environment returns as the coefficient to build an importance map. We also conducted experiments to investigate three major questions concerning frames' importance equality, the effectiveness of the importance map, and connections between importance maps from different IL models. The result shows that R2RISE successfully distinguishes important frames from the demonstrations.
\end{abstract}

\section{Introduction}
Recent advances in Imitation Learning (IL), which leverages external demonstration to reproduce the desired behaviours, demonstrate a promising performance in fields like 3D gameplay \cite{scheller2020sample}, robotics \cite{yu2018one}, and automatic driving \cite{codevilla2019exploring}. 
Most of the research on IL keep applying more complex Deep Neural Network (DNN) models, such as convolutional neural network (CNN) and generative adversarial network (GAN), to achieve greater performance under various condition while paying less attention to explaining what information the trained agents learned from the external demonstration. In this case, despite the success, IL methods are becoming increasingly unexplainable, and this problem remains an open challenge in both IL and Explainable Artificial Intelligence (XAI).


\begin{figure*}[t]
      \centering
      \includegraphics[width=0.95\linewidth]{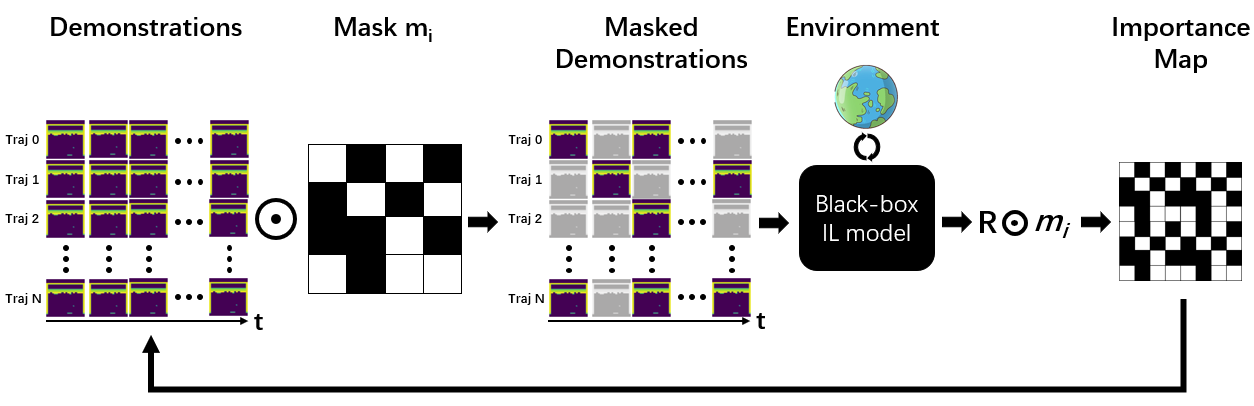}
      \caption{A diagrammatic representation of a single iteration of R2RISE. The input demonstrations are subject to element-wise multiplication (denotes as $\bigodot$) with a random mask which creates a masked demonstration, with greyed frames indicating those which are masked. Subsequently, the masked demonstration is used to train a black box IL model. The trained model interacts with the test environment to obtain returns, the mean of which is element-wise multiplied with the initial mask and accumulated to the existing importance map.}
      \label{fig_wf}
\end{figure*}

Currently, the number of existing research that combines XAI and IL is still limited, and they could be roughly categorized into two approaches to achieve better explainability, i.e., leveraging white-box models and analyzing the pixel-wise explainability via existing computer vision techniques. As for leveraging white-box models, most existing research substitutes the prevalent neural network architecture with other white-box models with intrinsic interpretability. For example, Leech ~\shortcite{leech2019explainable} proposed a learning framework that aggregates IL and logical automata to represent problems as compact finite state automata with human-interpretable logic states. Bewley et al. ~\shortcite{bewley2020modelling} modelled the behaviour policy of a trained black-box agent in the form of a decision tree by analyzing its input-output statistics. Zhang et al. ~\shortcite{zhang2021explainable} leveraged a hierarchical structure to explain the model's decision-making.
On the other hand, the research related to analyzing the pixel-wise explainability aims at the CNN structures in IL models that are widely used to capture features from image input. Referring to existing research in XAI and computer vision, the explainability of the model is commonly represented as heatmaps and analyzing the model's decision-making process from the heatmaps. For example, Pan et al. ~\shortcite{pan2020xgail} proposed a model-specific method called xGAIL that is based on Generative Adversarial Imitation Learning (GAIL) \cite{ho2016generative} and obtains local and global explanations for the passenger-seeking problem.

In fact, before xGAIL was proposed, the research community of IL investigated features in image inputs, but they did not highlight the significance of explainability. For example, Brown et al. ~\shortcite{brown2019extrapolating} used attention maps of the input image frames to validate the effectiveness of the learning process. De Haan et al. ~\shortcite{de2019causal} pointed out that the IL agent could learn wrong causal correlations between expert behaviours and irrelevant features in the input. These methods, including xGAIL, demonstrate what features in a single frame are significant for models to learn the desired behaviours. However, the above-mentioned methods fail to evaluate the importance of frames. Do the input image frames have identical importance? If not, how to distinguish frames' importance? 

To tackle these problems, we attempt to explain the input demonstrations as a whole by proposing a novel explaining method called R2RISE, which iteratively masks random frames in the demonstrations and evaluates the performance of the agents trained by the masked inputs. The intuition is that the input demonstrations are regarded as a single image, and frames in the demonstrations are regarded as pixels. In this case, existing XAI and computer vision methods could be directly applied to investigate the importance of frames instead of features in a single frame. R2RISE combines the existing methods RISE \cite{petsiuk2018rise} and ROAR \cite{hooker2019benchmark}, and achieves model-agnostic explanations for IL models with various architectures.

Our main contribution is summarized as follows:
1) We proposed a model-agnostic method to explain IL models;
2) We extended a novel perspective to explain IL with respect to the whole input dataset instead of a specific frame;
3) We investigated the connection between agents' overall performance and demonstration frames;

\section{ Preliminaries}

To better illustrate our approach, we first introduce the related existing literature in the field of XAI: RISE \cite{petsiuk2018rise} and ROAR \cite{hooker2019benchmark}. We then review an insightful method xGAIL \cite{pan2020xgail} that aggregates XAI with specific IL model GAIL \cite{ho2016generative}, and discuss its limitations. 

\subsection{Randomized Input Sampling for Explanation (RISE)}
Randomized Input Sampling for Explanation (RISE) is one of the state-of-the-art XAI methods proposed by Petsiuk et al. ~\shortcite{petsiuk2018rise} that explains black-box models. The attractive characteristics of RISE are its simplicity and generality. Unlike other popular XAI approaches, which calculate the gradient of image classification outputs, RISE probes the target model by randomly masking the input image and recording the probability result with respect to the target class. This process is repeated multiple times, and the recorded probabilities for each pixel are linearly combined to generate an importance map. This allows for the extraction of the most influential region in the input image for the target decision. RISE is also significant for explaining IL, as IL typically requires multiple demonstrations to train the model, which can be regarded as a single image. In this case, RISE can be used to explain which frames are important for policy training.

\subsection{RemOve And Retrain (ROAR)}
RemOve And Retrain (ROAR) was proposed by Hooker et al. ~\shortcite{hooker2019benchmark}, and its name concisely summarizes the working process of ROAR. By substituting some pixels estimated to be important with fixed uninformative values and then retraining a new model, ROAR achieves to evaluate feature importance for a wide range of models. The motivation of ROAR is that if the model demonstrates more sharp degradation in performance because of the removal, then we could conclude the proposed model is more accurate. The authors also argued that the retraining process is essential as machine learning models commonly assume that the training and test distribution is similar, and repeating the training several times could ensure a low variance in performance. 
The intuition of ROAR is valuable for explaining IL models, as training a single model to determine the importance of frames is risky, and the research community commonly uses ensemble methods to deal with the distribution shift. Retraining serval models under the same removal rate could improve the reliability of the importance along the demonstration trajectories. More importantly, since the conventional evaluation of IL problems is different from the classical CNN-involved XAI tasks, the performance of the trained IL model is represented as returns from the dynamic environment. It is improper to train the model once and feed image observations with fixed masks under such an environment.

\subsection{Explainable Generative Adversarial Imitation Learning (xGAIL)}
Pan et al. ~\shortcite{pan2020xgail} made the first attempt to explain one of the state-of-the-art models Generative Adversarial Imitation Learning (GAIL) \cite{ho2016generative}, and validated their method xGAIL on a passenger seeking problem. xGAIL was designed for problems that rely on spatial-temporal data, and both local and global explanations were obtained separately from a well-trained GAIL model. However, xGAIL's generality is severely limited to a specific problem and model. In addition, xGAIL, in fact, transforms the IL problem into an image classification problem by extracting and analyzing limited frames from abundant inputs. This could cause the absence of an overall explanation for the model's performance and generate explanations with bias as most of the information in the demonstration was filtered by the frame extraction process.

\section{R2RISE}

To overcome the above-mentioned limitations, we proposed a model-agnostic explanation method for imitation learning called R2RISE. R2RISE combines the merits of RISE and ROAR, and investigates the frames' importance with respect to the policy's overall performance.

We first review how RISE formulates the problem and distinguishes pixels' importance for the image classification problem. For a given image $\mathcal{I}$ with the size of $ H \times W $, RISE creates a random binary mask $m$ with the same size of $\mathcal{I}$, and does an element-wise multiplication between image $\mathcal{I}$ and mask $m$ (denoted as $\mathcal{I}\bigodot m$). The masked image then feeds into the black-box model (denoted as $f(\mathcal{I}\bigodot m)$). The importance of pixels is defined as the expected score over all possible masks $M = \{m_0,m_1,...,m_i\}$ conditioned on the event that pixel is observed (denoted as $M(\lambda)=1$, if the pixel is masked, then $M(\lambda)=0$), i.e. 
$S_{\mathcal{I},f}(\lambda) = \mathbb{E}_M [f(\mathcal{I}\bigodot m) | M(\lambda)=1].$
By rewriting the above equation as a summation over mask m and empirically estimating it using Monte Carlo sampling, the saliency map can be computed as a weighted sum of random masks and normalized by the expectation of M:
\begin{equation}
S_{\mathcal{I},f}(\lambda) \approx \frac{1}{\mathbb{E}[M] \cdot N} \sum_{i=1}^{N}f(\mathcal{I}\bigodot m_i) \cdot m_i(\lambda).
\end{equation}
Since RISE does not need any assumptions and information from the target model, RISE could be used to explain black-box models. The intuition behind RISE is that when $f(\mathcal{I}\bigodot m)$ is high, it indicates that the mask observes important pixels.
With similar intuition, Hooker et al. ~\shortcite{hooker2019benchmark} proposed ROAR to evaluate a feature importance. An ordered set of feature importance is estimated, and then they replace the top $l$ fraction of the ordered set with the corresponding channel mean, where $l$ is the pre-defined percentage of degradation level $l=[0,10,...,100]$. The major difference between RISE and ROAR is that ROAR retrains the model on the replaced dataset, while RISE only trains the model once.

Like most imitation learning methods, we assume the testing data has a similar distribution as training data, and the input demonstrations $\mathcal{D}_{n}$ are optimal. This could ensure evaluation fairness for a wide range of IL models. The demonstrations $\mathcal{D}_{n}$ consist of multiple trajectories, and each trajectory could be represented as either a sequence of state-action pairs or observations. In this work, we represent the trajectory as a sequence of state-action pairs, i.e. $\mathcal{D}_{n} = \{\tau_1,\tau_2,...,\tau_n\}$, where $\tau_{i \in [1,n]} = \{(s_1,a_1), (s_2,a_2),...,(s_t,a_t) \}$. The black-box imitation learning model trains a policy (denoted as $\pi_{\mathcal{D}_n}(a|s)$) on the input demonstrations $\mathcal{D}_{n}$, then interacts with the environment and obtains returns $R$ from the testing environment. For the finite horizon T, the expected return could be represented as the accumulation of the return at each time step, i.e. 
\begin{equation}
R(\pi_{\mathcal{D}_n}) = \mathbb{E}[\sum_{t=0}^T r_t | \pi_{\mathcal{D}_n}].
\end{equation}

\begin{figure*}[t]
     \centering
     \begin{subfigure}{0.3\textwidth}
         \centering
         \includegraphics[width=\textwidth]{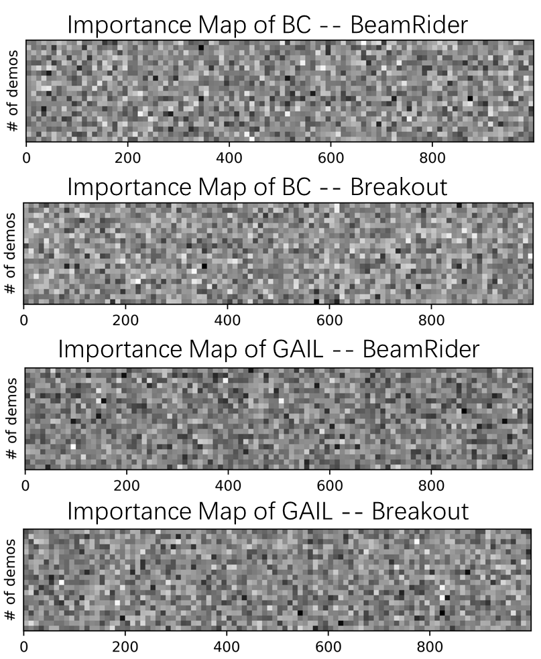}
         \caption{Importance maps generated by R2RISE.}
         \label{ims}
     \end{subfigure}
     \begin{subfigure}{0.65\textwidth}
         \centering
         \includegraphics[width=\textwidth]{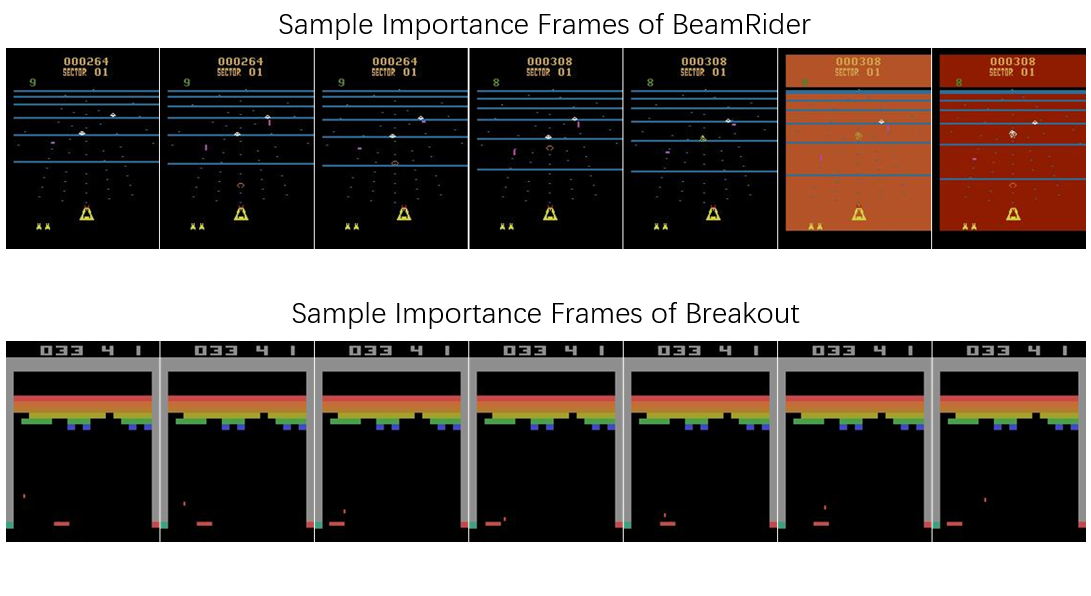}
         \caption{Important frames in BeamRider and Breakout.}
         \label{fms}
     \end{subfigure}
     \caption{Importance maps and the corresponding extracted sample frames that are recognized as important.}
\end{figure*}

The discussion we have so far motivated us to propose a frame-wise explanation method for IL called R2RISE. By regarding the demonstrations $\mathcal{D}$ as a single image, where the number of demonstrations is the image height $\mathcal{H}$, and the length of the demonstration is the image width $\mathcal{T}$, we could investigate the frame-wise importance based on the similar intuition of RISE. However, RISE could not be directly applied to IL since applying a fixed mask on a dynamic environment frame by frame is unreasonable, and IL methods are commonly evaluated by the interactions with the environment instead of feeding the policy network with another dataset. In this case, we aggregate ROAR with RISE and propose R2RISE. It hypothesises that the importance of each frame is not identical and iteratively removes random frames based on the predefined degradation level. The modified dataset $D_n = D \bigodot m_i$ is used to retrain an IL model. The retrained IL model then constantly interacts with the environment to obtain the accumulative return, and R2RISE finally compute the linear combination of the returns to obtain the saliency map (See Figure \ref{fig_wf}). Assuming the number of generated masks is $N$, and the return of each mask is the average return from $J$ rounds of interaction with the environment, the computation of the saliency map is similar to equation (1). To cater to the setting of IL, we substitute the $f(\mathcal{I}\bigodot m)$ in equation (1) with equation (2): 
\begin{align}
   S_{\mathcal{D}_n,f}(\lambda) & \approx \frac{1}{\mathbb{E}[M] \cdot N} \sum_{i=1}^{N} R(\pi_{\mathcal{D}_i}) \cdot m_i(\lambda) \\
   & = \frac{1}{\mathbb{E}[M] \cdot N} \sum_{i=1}^{N}  \mathbb{E}[\sum_{t=0}^T r_t | \pi_{\mathcal{D}_i}] \cdot m_i(\lambda) \\
   & = \frac{1}{\mathbb{E}[M] \cdot N \cdot J} \sum_{i=1}^{N}  
\sum_{j=0}^J \sum_{t=0}^T r_t  \cdot m_i(\lambda)
\end{align}
where $\mathcal{D}_i = \mathcal{D} \bigodot m_i$, and 
\[
m_i (\lambda) = 
\begin{cases}
0, & \mbox{if the pixel is masked}, \\
1, & \mbox{if the pixel is observed}.
\end{cases}
\]
As the formula presented, R2RISE also does not require any information from the IL models, such that R2RISE could be used as a model-agnostic method to explain IL.

To evaluate the effectiveness of R2RISE, we test two diverse IL methods: Behavioural Cloning (BC) \cite{bainFrameworkBehaviouralCloning1999} and Generative Adversarial Imitation Learning (GAIL) \cite{ho2016generative}. BC directly maps the states to actions from the input demonstration, and the control policy is obtained via supervised learning; GAIL, on the other hand, learns the policy through an iterative adversarial process between the generator G and discriminator D, where G is generating fake data distribution and D is differentiating the fake data distribution with the given expert distribution \cite{boyuan2022}. The ways BC and GAIL learn the policy are far different, and we wish to validate the generality of R2RISE from the diverse model selection.

\begin{algorithm}[tb]
    \caption{R2RISE}
    \label{alg:algorithm}
    \textbf{Input}: demonstrations $\mathcal{D}$, target IL model $f$\\
    \textbf{Parameter}: degradation level $l$, number of randomized masks $N$\\
    \textbf{Output}: an importance map $S_{\mathcal{D},f}$
    \begin{algorithmic}[1] 
        \STATE Initialize masks $M$ based on degradation level $l$ and number of randomized masks $N$.
        \STATE Initialize blank importance map $S_{\mathcal{I},f}$ with the same shape as $D$.
        \FOR{$m_i$ in M}
        \STATE Randomly initializes the model $f$.
        \STATE Obtain masked demonstrations $D_n = D \bigodot m_i$.
        \STATE Train model $f$ with the masked demonstrations $D_n$ and obtain policy $\pi_{D_n}$.
        \STATE Evaluate policy $\pi_{D_n}$ by interacting with environment and obtain average return $\Bar{R}$.
        \STATE Update importance map via element-wise addition, $S_{\mathcal{D},f} \leftarrow S_{\mathcal{D}_n,f} \bigoplus (\Bar{R} \bigodot m_i)$
        \ENDFOR
        \STATE \textbf{return} importance map $S_{\mathcal{I},f}$
    \end{algorithmic}
\end{algorithm}

\section{Experiment}
In this section, we conduct a series of experiments and address the following questions: (1) Is the importance between frames identical? (2) Can R2RISE distinguish the importance between frames? (3) Are there connections between the importance map obtained from different IL models?

\subsection{Setup}
We implement experiments with GPU NVIDIA Quadro RTX 5000, and two diverse IL models, BC and GAIL, are evaluated on two OpenAI Gym Atari tasks: Breakout and Beamrider \cite{brockman2016openai}.

Similar to recent IL methods, we leverage the proximal policy optimization (PPO) \cite{schulman2017proximal} algorithm from the OpenAI baselines \cite{dhariwal2017openai}, utilizing default parameters and reward function, to generate expert demonstrations. The PPO training process is checkpointed every 20 steps, and the observations with the size of $84 \times 84 \times 3$ and actions between the PPO agents and the task environment are recorded as "trajectories." These trajectories, generated from checkpoint 1400, serve as expert demonstrations. To avoid the ``causal confusion" problem (models build wrong causal relationships with irrelevant patterns) \cite{dehaanCausalConfusionImitation2019} and ensure the fairness of our evaluation, we mask the indicators (such as scoring broad) in frames and ensure the same demonstrations as input for different IL models.

Regarding the parameter setting, we generate 20 trajectories with a fixed length of 1000 for each IL model. Five random seeds and five levels of percentage degradations $l = [10,30,50,70,90]$ are pre-defined for evaluation. To assess each given random seed and percentage degradation, we propose to retrain 100 models with 100 randomized masks. Each mask contains 20*100 grids, which means that every single trajectory is cut into 100 snippets, and each snippet assigns the same importance to 10 frames. The retrained model is tested from 20 trials, and the average return of the trials, multiplied element-wise with the random mask, is added to the final importance map.

\begin{figure}[ht]
     \centering
     \begin{subfigure}{0.45\textwidth}
         \centering
         \includegraphics[width=\textwidth]{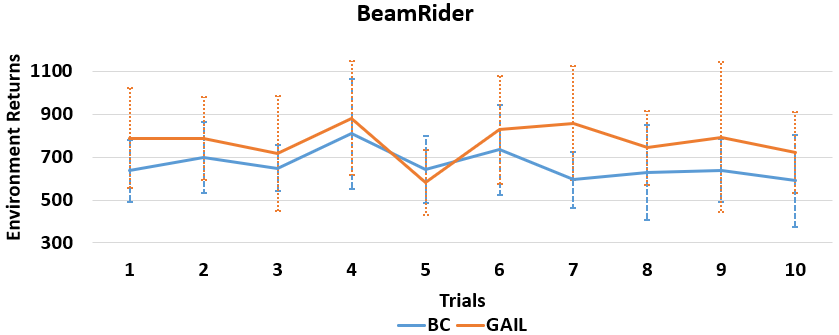}
     \end{subfigure}
     \begin{subfigure}{0.45\textwidth}
         \centering
         \includegraphics[width=\textwidth]{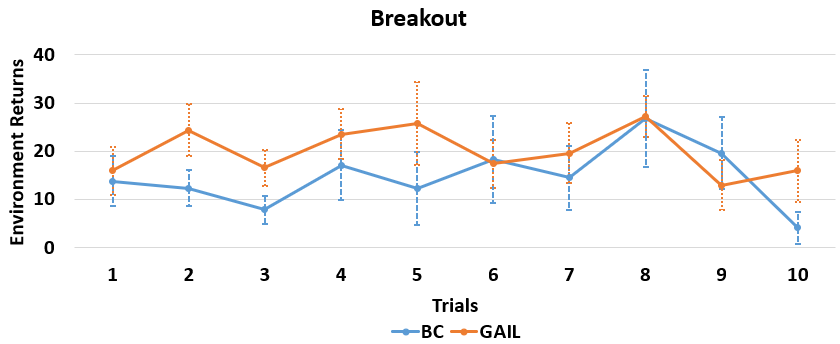}
     \end{subfigure}
     \caption{Deviations in policy performance of 10 models trained by 10 different randomized masks.}
     \label{exp1_10}
\end{figure}

\subsection{Is the importance between frames identical? }

Remember that we hypothesize that the importance of frames is different, so we validate this hypothesis by applying several randomized masks on the same demonstration and comparing the performance of the trained model. If the outcomes present noticeable deviations, then it can be inferred that the contribution between frames varies. To further this idea, we divide each trajectory into ten segments of equal length and randomly assign either a value of 0 or 1 to each segment. Regions assigned 0 are removed, and then the preprocessed demonstrations are used as input to train the relevant models.
This process iterates ten times, and we get Figure \ref{exp1_10}. Here, the x-axis is the number of attempts, the y-axis is the environment returns, and the error bars indicate the standard deviation of the policy performance. 

From Figure \ref{exp1_10}, we can observe the performances deviate from model to model. The best model achieves far better performance than the worst one, which indicates that the importance of frames is not identical. Similarly, the importance maps generated by R2RISE also suggest disparities between frames (see Figure \ref{ims}). The x-axis is the trajectory's length, the y-axis represents the trajectories, and the grayscale denotes the importance. The whiter the grid is, the more important the grid is. Figure \ref{fms} shows the extracted frames that are recognized as the most important components in the demonstrations. For the task of BeamRider, models would put more weight on destroying enemy flights, whereas for the task of Breakout, models pay more attention to the rebounding process of the upper blocks, sidewalls, and paddle, which resembles the strategy we might use for these games.


\begin{figure}[t]
     \centering
     \begin{subfigure}{0.45\textwidth}
         \centering
         \includegraphics[width=\textwidth]{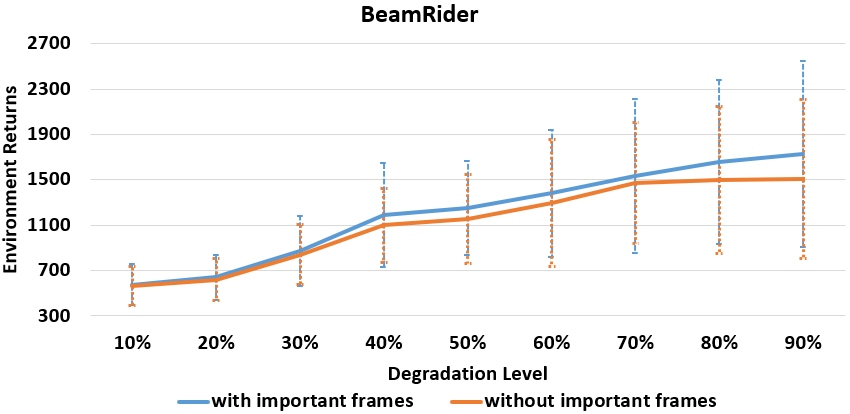}
         \caption{Policy performance changes in BeamRider.}
         \label{vd_beam}
     \end{subfigure}
     \begin{subfigure}{0.45\textwidth}
         \centering
         \includegraphics[width=\textwidth]{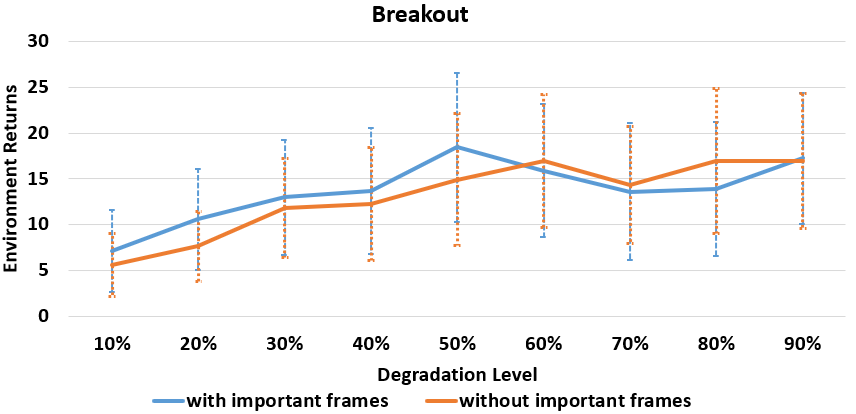}
         \caption{Policy performance changes in Breakout.}
         \label{vd_breakout}
     \end{subfigure}
     \caption{Validations of the effectiveness of R2RISE. The lines and the error bars represent the mean performance and standard deviation of the model trained from a certain percentage of the most important (or least important) frames. }
     \label{vd}
\end{figure}

\subsection{Can R2RISE distinguish the importance between frames? }

This section investigates the effectiveness of R2RISE. According to the abovementioned hypothesis validation, we obtain a map indicating the importance of frames. However, the quality of the generated maps needs to be properly evaluated. In this case, we use similar causal metrics in \cite{petsiuk2018rise}, insertion and deletion, where the availability of the 'cause' will significantly influence the model's decision-making and performance. Under the scenario of image classification tasks, deleting the causal pixels will lead to a sharp drop in accuracy if the model gets well explained. In our experiment, we leverage similar intuition and expect the removal of the important frames would lead to a worse performance while limiting the amount of input data to be the same. To achieve this, we transform the generated importance map into a mask by setting up a threshold and replacing the map with either 1s or 0s, depending on the threshold.

Figure \ref{vd} shows the changes in policy performance using different percentages of the most important (or least important) frames. The x-axis is the percentage of data used to train the model, and the y-axis is the environment returns. The solid lines are the average returns from 20 trials using the same transformed mask and demonstrations, the error bar is the standard deviation of the 20 trials. From Figure \ref{vd}, we can observe that the models trained with the most important frames perform well when the input data is relatively limited for both tasks, which meets our expectations. For task BeamRider (see Figure \ref{vd_beam}), the model with important frames always performs better than the model trained with the least important frames. The performance deviation at the beginning of the figure is relatively small, we think the reason is that the model is more sensitive to the amount of data than to the availability of the important frames. From the end of this figure, it can be seen that the performance deviation increases, which indicates that the missing top important frames significantly limit the upper bound of the model's performance. For task Breakout (see Figure \ref{vd_breakout}), it can be seen that the model with important frames significantly outperforms the model without important frames until more than 60\% of the total demonstrations are fed. When more than 60\% of the data is given, the model's performance starts to fluctuate. We suspect this is due to the fact that 50\% of the initial input is sufficient to train the policy, and with more ordinary or redundant frames added to the dataset, the policy performance is negatively influenced.

\begin{figure}[t]
     \centering
     \begin{subfigure}{0.45\textwidth}
         \centering
         \includegraphics[width=\textwidth]{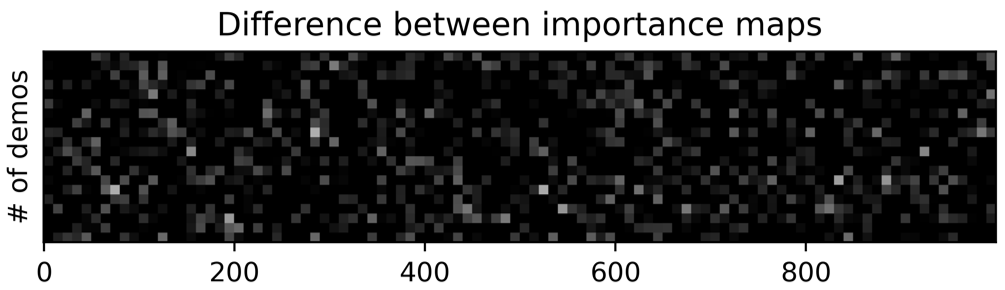}
         \caption{Deviation between the importance maps in BeamRider.}
         \label{dev_beam}
     \end{subfigure}
     \begin{subfigure}{0.45\textwidth}
         \centering
         \includegraphics[width=\textwidth]{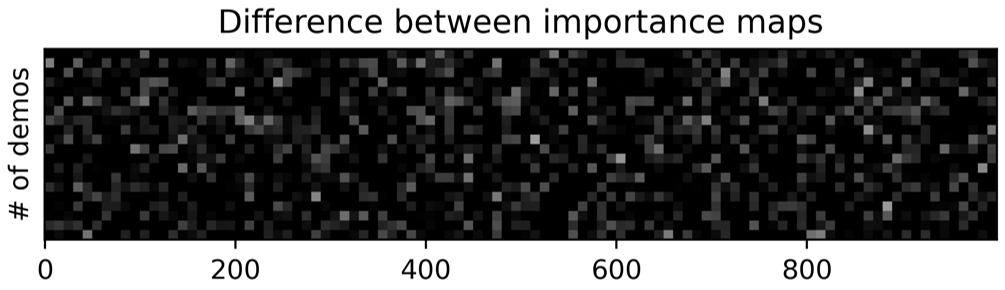}
         \caption{Deviation between the importance maps in Breakout.}
         \label{dev_breakout}
     \end{subfigure}\
     \caption{Deviation between the importance maps of BC and GAIL.}
     \label{dev}
\end{figure}

\subsection{Are there connections between the importance map obtained from different IL models?}
\begin{figure}[t]
      \centering
      \includegraphics[width=0.45\textwidth]{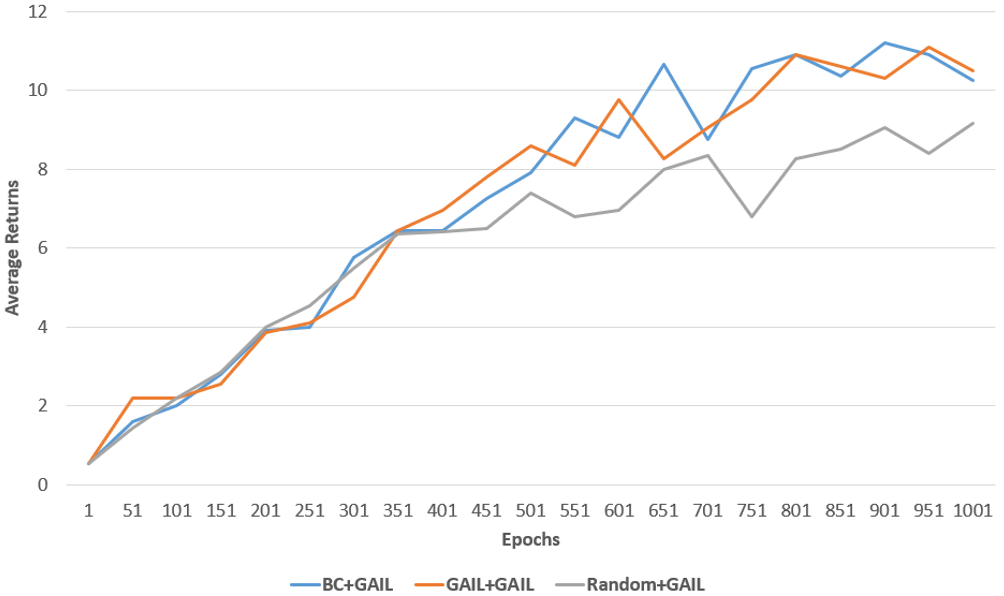}
      \caption{Average reward learning curves of GAIL trained with different masks. The blue, orange and grey lines are trained with the masks extracted from the importance map obtained using BC, GAIL and random, respectively. }
      \label{fig_training}
\end{figure}

In addition to exploring the explainability within a single IL model, we also attempt to examine the connections between IL models. In this section, we investigate the question: does the importance map obtained by one model have connections to another model? To this end, we propose two approaches to explore intrinsic connections. The first attempt directly compares the importance maps by projecting the values into the same range and calculating element-wise deviation (see Figure \ref{dev}). The larger the deviation is, the whiter the output image should be. From Figure \ref{dev}, we can observe that most grids are close to black, which indicates the assigned importance for these grids are similar. 
The second attempt involves generating a mask from an importance map generated by one model and applying that mask to the same demonstrations to train another model. The underlying assumption is that if there are connections between models' importance maps, the important frames identified by one model should work well on another model, leading to improved performance compared to those without the frames. Figure \ref{fig_training} displays the average returns of GAIL using three types of masks: the blue, orange and grey lines correspond to models trained with masks extracted from the importance map obtained using BC, GAIL and random, respectively. It can be seen in Figure \ref{fig_training} that the model trained with BC's mask demonstrates similar policy performance to the model trained with GAIL's mask, and both models outperform the model using a randomized mask under the same training epoch. This confirms our expectation that the importance frames recognized by BC could also be employed to train model GAIL with considerable performance. This could prove useful when the target model is time-consuming, and one could use a more time-efficient method like BC to obtain an alternative importance map.

\section{Limitations}
Several intriguing challenges await further exploration. Although we were able to extract promising explanations through a large number of masks and validations of the obtained policy, the robustness of the framework could be further improved through ensemble methods. As the framework currently relies on a single trial to train the model, ensembling IL models could reduce performance variance for each given mask. In addition to robustness, computation intensiveness is another limitation. The time taken to obtain a satisfactory explanation is closely linked to the time spent on training the target model once. If the target model requires days to train, it would not be practical to retrain hundreds of times. Improving time efficiency while preserving the model-agnostic property remains an open challenge for R2RISE. Investigating the relationship between global explanations and the frames that are recognized as important is another interesting future direction. Although we observed similar patterns in the extracted frames from different trials and models, it is still unsafe to claim these patterns could be the global explanations for a specific task. Further research is needed to provide theoretical proof for the connections.

\section{Conclusion}
This paper introduced a model-agnostic explaining framework for imitation learning called R2RISE. R2RISE distinguishes the frames' importance in relation to the overall policy performance. It iteratively applies numerous randomized masks on the demonstrations and retrains the black-box IL model based on the masked demonstration. Evaluation of the obtained policy is conducted in a manner similar to most IL methods, where the policy is evaluated by the accumulated returns from the environment, and we leverage the accumulated returns as a coefficient to multiply with the mask to obtain the importance map of the frames. Experiments have shown that R2RISE can successfully distinguish important frames from the demonstrations, thus providing insight into which frames contribute to better performance.

\clearpage
\bibliographystyle{named}
\bibliography{ijcai22}

\end{document}